\documentclass[runningheads]{llncs}

\usepackage[T1]{fontenc}
\usepackage{graphicx}
\usepackage{booktabs}
\usepackage{multirow}
\usepackage{caption}
\usepackage{subcaption}

\usepackage[accsupp]{axessibility}  % Improves PDF readability for those with disabilities.

\usepackage{hyperref}

\usepackage{orcidlink}

\begin{document}

\title{OneDiff: A Generalist Model for Image Difference Captioning} 


\author{Erdong Hu\inst{1,2} \and
Longteng Guo\inst{1} \and
Tongtian Yue\inst{1,2} \and
Zijia Zhao\inst{1,2} \and
Shuning Xue\inst{1,2} \and
Jing Liu\inst{1,2}\thanks{Corresponding authors.}}

\authorrunning{E.~Hu et al.}

\institute{Institute of Automation, Chinese Academy of Sciences, China \and
School of Artificial Intelligence, University of Chinese Academy of Sciences, China\\
\email{\{huerdong2022, yuetongtian2022, zhaozijia2021, xueshuning2021\}@ia.ac.cn,}\\
\email{\{longteng.guo, jliu\}@nlpr.ia.ac.cn}}
% \institute{Institute of Automation, Chinese Academy of Sciences, China \\
% \email{\left\{huerdong2022, yuetongtian2022, zhaozijia2021, xueshuning2021\right\}@ia.ac.cn,} \\
% \email{\left\{longteng.guo, jliu\right\}@nlpr.ia.ac.cn} \and
% School of Artificial Intelligence, University of Chinese Academy of Sciences, China}

\maketitle

\begin{abstract}
In computer vision, Image Difference Captioning (IDC) is crucial for accurately describing variations between closely related images. Traditional IDC methods often rely on specialist models, which restrict their applicability across varied contexts. This paper introduces the OneDiff model, a novel generalist approach that utilizes a robust vision-language model architecture, integrating a siamese image encoder with a Visual Delta Module. This innovative configuration allows for the precise detection and articulation of fine-grained differences between image pairs. OneDiff is trained through a dual-phase strategy, encompassing Coupled Sample Training and multi-task learning across a diverse array of data types, supported by our newly developed DiffCap Dataset. This dataset merges real-world and synthetic data, enhancing the training process and bolstering the model’s robustness. Extensive testing on diverse IDC benchmarks, such as Spot-the-Diff, Image-Editing-Request, and Birds-to-Words, shows that OneDiff consistently outperforms existing state-of-the-art models in accuracy and adaptability, achieving improvements of up to 97\% CIDEr points in average. 
By setting a new benchmark in IDC, OneDiff paves the way for more versatile and effective applications in detecting and describing visual differences. The code, models, and data will be made publicly available.


\end{abstract}

\section{Introduction}
\label{sec:intro}

Humans possess a remarkable ability to detect differences within visual scenes, a skill critical for myriad real-world applications such as change detection in surveillance footage, anomaly detection in medical images, and more. This human capability has inspired the development of Image Difference Captioning (IDC), a task in computer vision aimed at algorithmically describing discrepancies between two similar images. IDC is pivotal for precise difference detection required in fields like medical diagnostics, environmental monitoring, and manufacturing quality control. Additionally, IDC holds potential for aiding in tasks like distinguishing closely related species, describing lesions in medical images, monitoring changes in media assets, and differentiating actions in videos \cite{10562229,10183357}.

Despite the clear importance and growing interest in IDC, current methodologies largely focus on specialist models tailored for specific scenarios. These specialist approaches typically utilize intricate attention mechanisms and advanced image processing techniques to highlight subtle differences. However, they often lack the flexibility needed to generalize across different datsets and tasks, which limits their practical applicability and scalability. 
% These models usually perform well under controlled conditions or within the narrow domains for which they were designed, but struggle when applied to new or varied datasets.

\begin{figure}[tb]
    \centering
    \includegraphics[width=\linewidth]{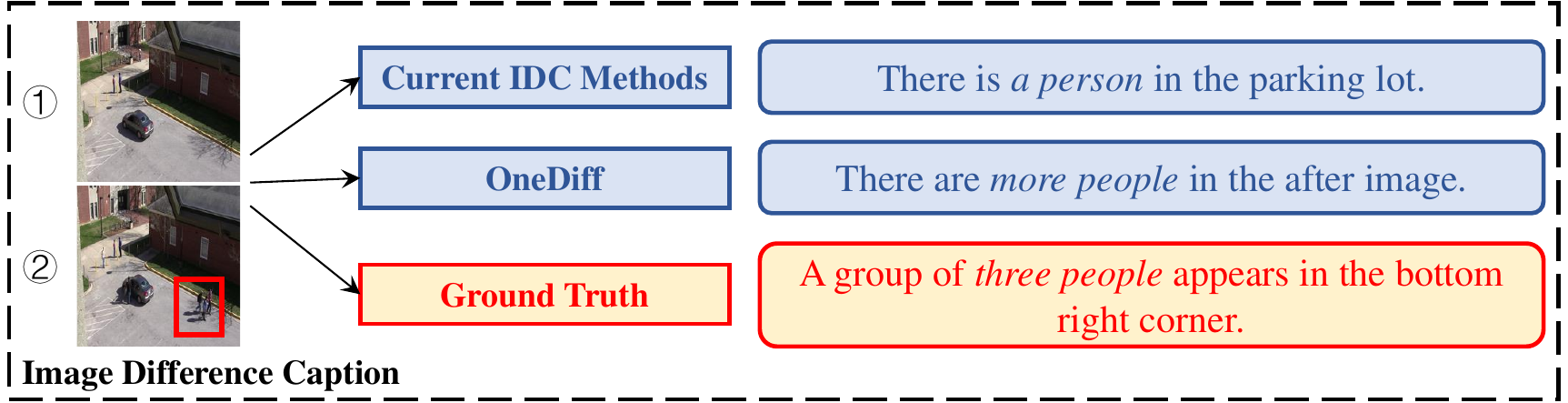}
    \caption{Illustration of Image Difference Captioning tasks.}
    \label{fig:intro}
\end{figure}

This paper aims to address these limitations by proposing a generalist model for IDC. By harnessing the power of large Vision-Language Models (VLMs), such as those demonstrated by GPT-4 \cite{achiam2023gpt}, Gemini \cite{team2023gemini} and LLaVA \cite{llava}, which have shown significant capabilities in aligning visual and textual features. These VLMs, however, still fall short in tasks requiring meticulous detail recognition between image pairs, such as IDC. We intend to develop a model to harness and extend these capabilities to not only capture the intricate details necessary for accurate IDC but also to thrive across diverse scenarios.

The development of such a model presents two primary challenges: capturing fine details and the scarcity of data. IDC requires an acute perception of fine-grained differences, often subtle and specific to the context, like variations in bird plumage or slight changes in manufactured parts. Additionally, high-quality annotations for IDC are costly and labor-intensive, necessitating detailed comparison and description by human annotators, which limits the availability of large-scale datasets.

To address these challenges, we introduce the \textbf{OneDiff model}, a novel LVLM specifically designed to harness the intricate details present in image pairs and trained on a vast and varied collection of IDC datasets. This section of the paper delineates our methodology, beginning with the architecture of the model and moving on to its innovative training approach.
At the core of OneDiff lies a hybrid architecture that synergizes a siamese image encoder with a large language model (LLM), enabling efficient and precise capture of visual discrepancies. The siamese image encoder processes each image of the pair independently, subsequently fusing their representations. Key to our architecture is the introduction of \textbf{Visual Delta Module}. It includes learnable query tokens that probe the multi-level encoded features of both images, from lower to higher layers, effectively capturing their differences. This mechanism not only enhances the model’s ability to discern fine-grained details but also leverages the powerful generative capabilities of LLMs to produce descriptive and contextually relevant captions.

Our training strategy is structured in two phases to optimize the model’s performance and adaptability across various IDC tasks. In the first phase, we focus on aligning visual and linguistic modalities through a simple yet effective technique called  \textbf{Coupled Sample Training}. It dynamically combines different image pairs into pseudo dual-image samples, providing a rich training ground for the model to learn complex pair-wise image-text alignments.
The second phase involves multi-task training utilizing a diverse array of IDC data. Here, each dataset is standardized into a unified instruct-tuning format, guided by task-specific prompts.

To effectively train our generalist IDC model, we create \textbf{DiffCap Dataset}, an open-domain, large-scale dataset comprising three distinct types of IDC data: aggregated task-specific collections and two forms of synthetic data that we developed. First, we aggregate existing IDC datasets to form a solid base of varied real-world scenarios. Second, we expand our dataset with open-domain real image pairs, identified through image library searches and equipped with GPT-assisted autogenerated captions to increase coverage and context variability. Finally, we enhance DiffCap with synthetic images from InstructPix2Pix \cite{brooks2023instructpix2pix} dataset, which are specifically designed to include subtle and complex changes. These synthetic images are paired with change summaries generated by GPT-3 \cite{brown2020language}, addressing the dual challenges of training data scarcity and limited manipulation variety in conventional IDC datasets. This strategic combination allows our model to adapt and perform robustly across a wide array of scenarios.

By integrating these advanced methodologies, OneDiff effectively overcomes the longstanding challenges of data scarcity and detail capture in IDC. 
Extensive experiments on diverse IDC benchmarks such as Spot-the-Diff, CLEVR-Change, and Birds-to-Words reveal that our OneDiff model significantly outperforms current state-of-the-art methods in essential metrics without the need for benchmark-specific tuning, affirming its robust generalization capabilities.

The principal contributions of this work include:
\begin{itemize}
\item We propose OneDiff, a new generalist model for Image Difference Captioning that incorporates Visual Delta Module for precise detail capture and Coupled Sample Training for enhanced training robustness. This framework allows the model to excel across varied IDC tasks.

\item We introduce DiffCap, a rich dataset that merges real-world and synthetic data to address the scarcity of diverse and complex training samples.

\item The OneDiff model demonstrates outstanding performance across multiple IDC scenarios, setting new benchmarks for flexibility and efficiency in the field without the need for specific adaptations for different tasks.
\end{itemize}

\section{Related Work}

\subsection{Image Difference Captioning}

Image difference captioning (IDC) emphasizes subtle variations between images by focusing on changes rather than common elements. The pioneering approach in this field, Spot-the-Diff \cite{jhamtani2018learning}, introduces an LSTM-based \cite{sak2014long} network to detect and model potential change clusters by analyzing differences at the pixel level between two images. This method, however, is susceptible to noise and geometric transformations. In contrast, DUDA \cite{park2019robust} enhances robustness by calculating image differences at the semantic level using CNNs, effectively handling minor global changes.

Subsequent developments in IDC include M-VAM \cite{shi2020finding} and VACC \cite{kim2021agnostic}, which incorporate a viewpoint encoder to address differences in perspective. VARD \cite{tu2023adaptive} further advances this by introducing a network that offers viewpoint invariant representations to capture changes explicitly. In addition, \cite{sun2022bidirectional} improve change localization through bidirectional encoding, and NCT \cite{tu2023neighborhood} enhances feature aggregation using a Transformer. These methods typically leverage specific properties of image modality relevant to their benchmarks, such as nearly identical views in Spot-the-Diff \cite{jhamtani2018learning} or controlled variables like object and change types (e.g., color, texture, addition, and removal) in synthetic scenes of CLEVR \cite{johnson2017clevr}.

Recent innovations in IDC, such as IDC-PCL \cite{yao2022image} and CLIP4IDC \cite{guo2022clip4idc}, employ BERT-like training techniques to enhance the modeling of language for difference captioning. These approaches have achieved state-of-the-art results, showcasing significant advancements in the field of IDC.

\subsection{Large Vision-Language Models}

The fusion of computer vision and natural language processing has led to the development of Vision-Language Models (VLMs), which combine visual and linguistic elements to foster cross-modal comprehension and reasoning. This synergy has been crucial in enhancing tasks that necessitate both visual perception and language interpretation, as highlighted by models trained with various datasets aimed at enhancing understanding and reasoning \cite{antol2015vqa,hudson2019gqa}. Notable breakthroughs, including CLIP, have significantly narrowed the divide between language and vision tasks, demonstrating the viability of cross-modal applications. Recent trends indicate an increased focus on harnessing the powerful capabilities of Large Language Models (LLMs) in the VLM domain. Since LLMs can only perceive text, bridging the gap between natural language and vision information is necessary for the development of large vision-language models.

A common and feasible solution is to leverage a group of learnable query tokens to extract information in a query-based manner, which first has been implemented in BLIP-2 \cite{li2023blip}, and subsequently inherited by a variety of work. Such Q-Former style approaches compress visual tokens into a smaller number of representation vectors. In contrast, some methods simply use a MLP-based interface to bridge the modality gap. For example, LLaVA \cite{llava} series adopts one/two linear MLP to project visual tokens and align the feature dimension with word embeddings. 

As another line, feature-level fusion inserts extra modules that enable deep interaction and fusion between text features and visual features. For example, Flamingo \cite{alayrac2022flamingo} inserts extra cross-attention layers between frozen Transformer layers of LLMs, thereby augmenting language features with external visual cues. Similarly, CogVLM \cite{wang2023cogvlm} plugs in a visual expert module in each Transformer layer to enable dual interaction and fusion between vision and language features. For better performance, the QKV weight matrix of the introduced module is initialized from the pre-trained LLM. Similarly, LLaMA-Adapter \cite{zhang2023llama} introduces learnable prompts into Transformer layers. These prompts are first embedded with visual knowledge and then concatenated with text features as prefixes.

\begin{figure}[tb]
  \centering
  \includegraphics[width=1\linewidth]{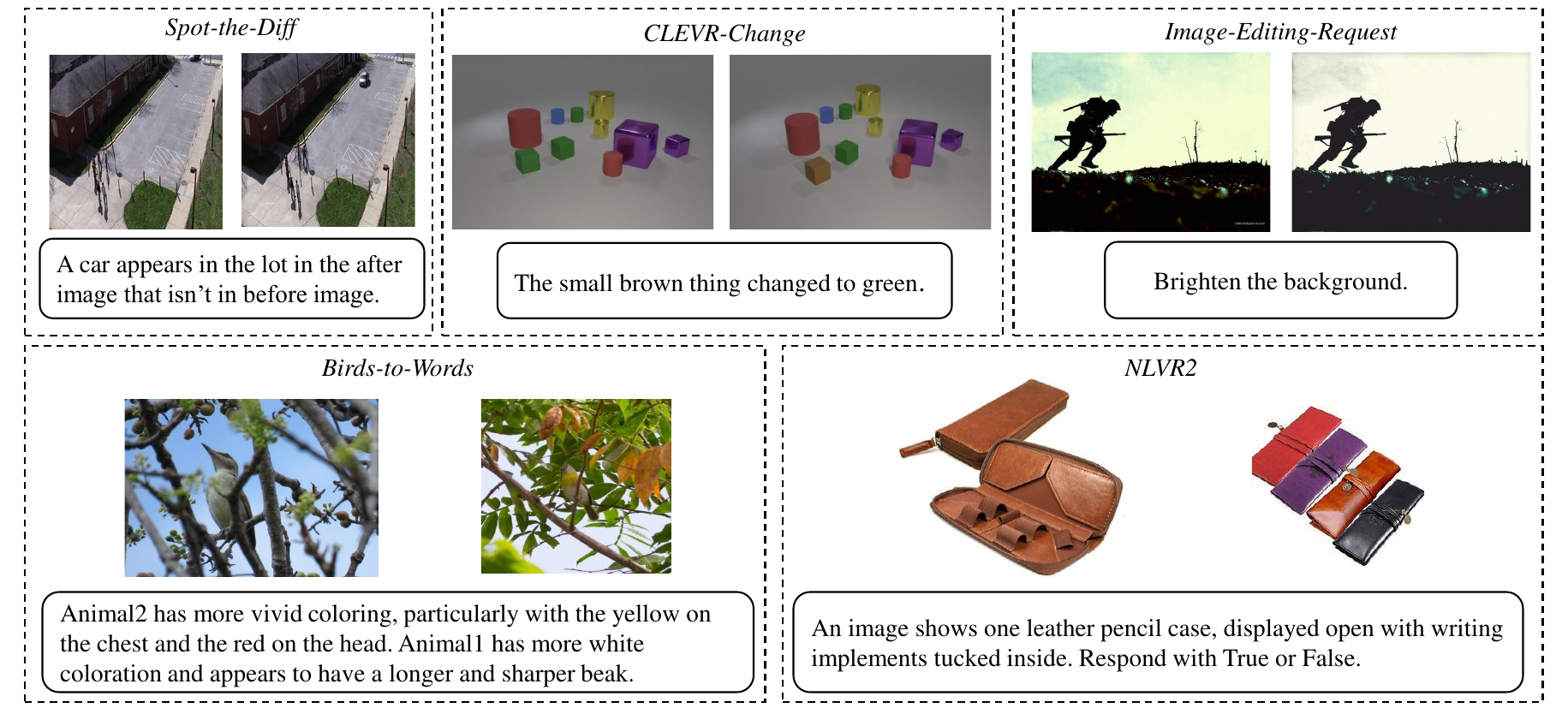}
  \caption{Data components and examples in the task-specific data collection of our DiffCap dataset.
  }
  \label{fig:coll}
\end{figure}

\section{DiffCap: A Multi-Task Dataset for IDC}

The effectiveness of Vision-Language Model (VLM), particularly in tasks like Image Difference Captioning (IDC), is heavily dependent on the quality and diversity of its training data. To address the challenges of data scarcity and variability, we develop the DiffCap Dataset. This comprehensive dataset is tailored to enhance robustness and generalization of our OneDiff model. In this section, we describe the design, composition, and key statistics of our DiffCap Dataset. Specifically, DiffCap is meticulously designed to encompass a wide range of scenarios, aiming to address common challenges encountered in practical applications. It integrates both realistic and synthetic data to create a balanced and extensive collection. This strategic design ensures that the model is trained not only on frequent, common scenarios of image differences but also exposed to rare and intricate changes that are often overlooked in other datasets.

\subsection{Task-Specific Data Collection}

We firstly curate data from existing datasets that focus on identifying differences between image pairs, as is shown in \ref{fig:coll}. These include benchmarks like Spot-the-Diff \cite{jhamtani2018learning} and CLEVR-Change, which are directly compatible with our IDC tasks. Additionally, the Birds-to-Words dataset, which is tailored for describing differences between birds, and the Image-Editing-Request dataset, which provides instructions for editing images, are also included. The NLVR2 dataset, known for its dual-image inputs, further enhances the model's ability to capture intricate details in image pairs.

This collected data has been converted into a unified instruction-following format. Each entry includes a pair of input images, a task-specific natural language prompt as a query instruction, and a difference caption as the ground-truth response. The varied domains and styles of these datasets underscore their potential as benchmarks for our research. Detailed descriptions of these datasets are provided in the Experiments section.

\subsection{Automatic Data Construction with Real Images}

To reflect real-world scenarios and enable broader applications, priority has been given to acquiring image pairs from real images for IDC data generation. 

\begin{figure*}[tb]
    \centering
    \includegraphics[width=\linewidth]{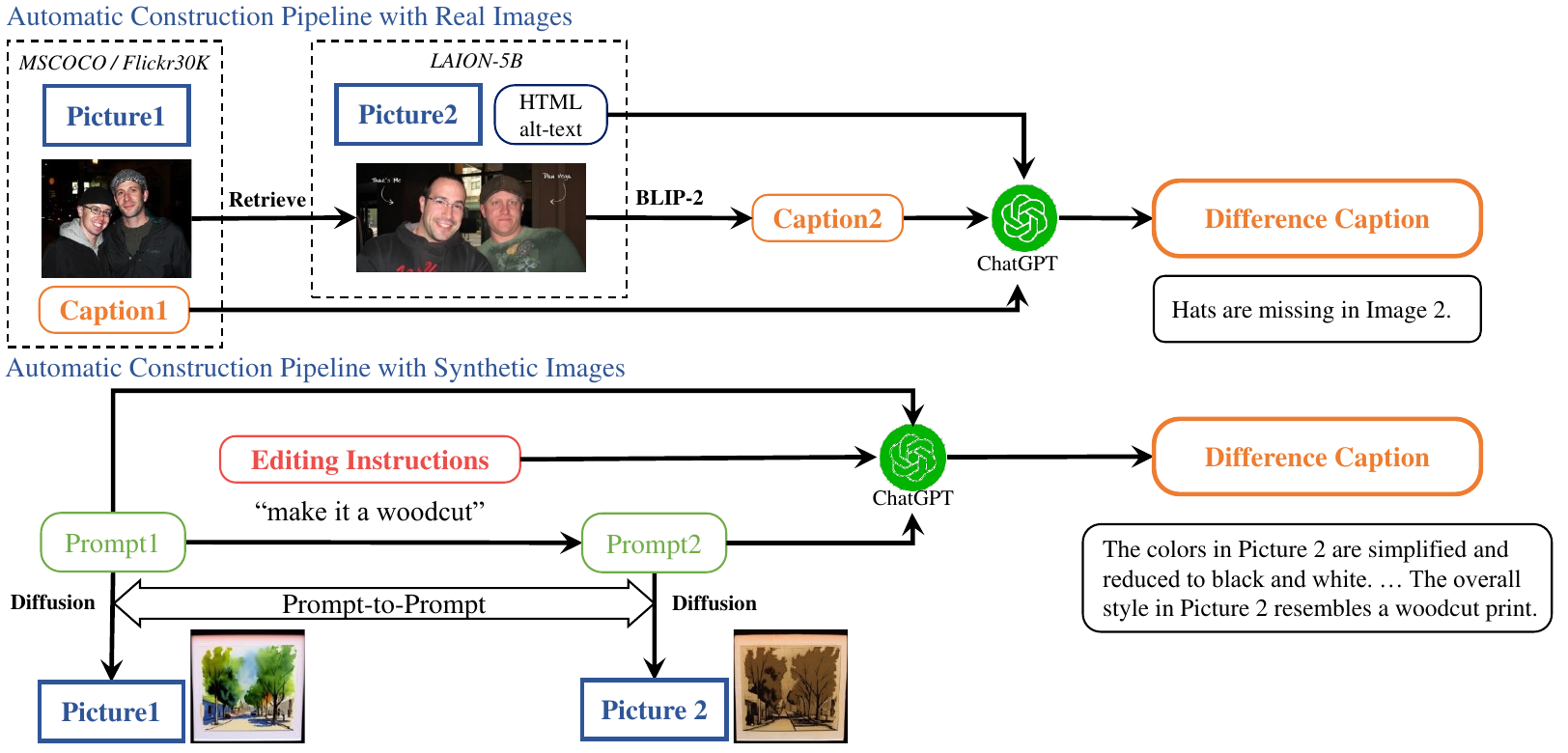}
    \caption{Automatic data construction pipeline of our DiffCap dataset.}
    \label{fig:gen}
\end{figure*}

\paragraph{Real Images Pairs Collection.}
We implement a retrieval-based approach to collect similar images from a large image database, LAION-5B, with source images from MSCOCO \cite{lin2014microsoft} and Flickr30K \cite{plummer2015flickr30k} datasets. Based on the cosine similarity between two CLIP image features. We filter out image pairs with a similarity score below 0.75 due to high disparity and above 0.94 due to near identity. In addition, we exclude image pairs with highly similar textual information to ensure the visibility of granular visual differences during annotation.

\paragraph{Difference Caption Annotation.}
Following the collection of similar image pairs, we gather necessary textual information for annotation by downstream large language models. Original images sourced from COCO \cite{lin2014microsoft} and Flickr30K \cite{plummer2015flickr30k} include manually annotated captions, while retrieved images from LAION-5B feature associated HTML alt-text. Despite the filtering for visual-textual similarity, the alt-text quality from LAION-5B often remains low for use as captions. We employed BLIP-2 to generate higher-quality captions for the retrieved images. To alleviate possible hallucinations introduced by BLIP-2, We calculate the image-text similarity between retrieved images and their generated captions using CLIP, and filter for data with scores over 40. These captions, along with the HTML alt-texts, are fed into ChatGPT to generate descriptive captions that articulate the differences between image pairs. 
% Before conducting large-scale automatic annotation, we optimize the prompt and examples by randomly selecting 3 groups of 100 data entries for annotation and manual check. If 90 or more cases pass the standard in all 3 trials, we deem the GPT-generated data to be of relatively assured quality.

\subsection{Automatic Data Construction with Synthetic Images}
While image pairs retrieved by our methods generally display high-level semantic differences, they still exhibit significant low-level visual discrepancies. To better capture these fine-grained details, we have included synthetic images with subtle differences in our dataset to improve generalization capabilities by data synergy. 

\paragraph{Synthetic Image Pairs Collection.}
Our method incorporates the generated paired image editing examples from the InstructPix2Pix \cite{brooks2023instructpix2pix} framework. InstructPix2Pix uses StableDiffusion in combination with Prompt-to-Prompt to generate pairs of synthetic images from pairs of closely related captions,  ensuring each synthetic image pair contains subtle, yet significant, differences relevant to IDC tasks. 
 
\paragraph{Difference Caption Annotation.}
Given that the synthetic image pairs are derived from text prompts guiding the Stable Diffusion and Prompt-to-Prompt processes, these prompts inherently contain the modifications made to the original image. However, traditional image editing annotations often focus only on the outcome, missing the context of the original content. 
To bridge this gap, we employ ChatGPT to generate difference captions that comprehensively describe the changes between the original and edited images. We prompt ChatGPT with the original image prompt, the edited image prompt, and the specific editing instruction to produce a difference caption. 
For instance, if the original prompt describes "a serene lake scene" and the editing prompt modifies it to "a serene lake scene with a kayak," the direct instruction might simply be "add a kayak." 
% However, for our IDC tasks, it is more informative to generate a difference caption like "a kayak has been added to the serene lake scene," which explicitly points out the addition within the unchanged context of the lake scene.

\subsection{Dataset Statistics}

We develop automatic pipelines to generate image difference captioning data from both realistic and synthetic images for our DiffCap dataset. We have selected 53k pairs of real images, each annotated with difference captions averaging 12.2 words in length. Our synthetic part includes 139k pairs of images and corresponding annotations with an average length of 15.0 words. As supplement, we collect and sample data from existing benchmarks with dual-image inputs, forming a data mixture of 170k image pairs. Aggregating data from above three parts, we have compiled our dataset DiffCap which totals 363k entries detailed in our supplementary materials.
% \ref{data:diff}.

% \begin{table}[t]
% \caption{Statistics of our DiffCap dataset. `Real" refers to our generated data from real images, and `Synthetic" refers to our generated data from synthetic images.}
% \centering
% \begin{tabular}{llrr} 
% \toprule
% Component & Sub-Component & \#Image-Pairs & \#Annotations \\
%  \midrule
% \multirow{5}{*}{Collected} & Spot-the-Diff \cite{jhamtani2018learning} & 11K & 21K\\
%  & CLEVR-Change & 8K & 44K\\
%  & Image-Editing-Request & 3.4K & 4.2K\\
%  & Birds-to-Words & 14K & 14K\\
%  & NLVR2 & 86K & 86K\\
% \midrule
% \multirow{2}{*}{Generated} & Real & 54K & 54K\\
%  & Synthetic & 139K & 139K\\
% \midrule
% Total &  & 316K & 363K\\
% \bottomrule
% \end{tabular}
% \label{data:diff}
% \end{table}

\section{OneDiff: A Generalist Model for IDC}

OneDiff is a novel vision-language model with the Visual Delta Module specially designed to capture more comprehensive vision features. We generate both realistic and synthetic data for training to improve the capability of extracting detailed differences between image pairs. Our method endows traditional vision-language models with competency to tackle tasks with multi-image inputs and exhibits strong adaptability and scalability.

\begin{figure}[tb]
    \centering
    \includegraphics[width=\linewidth]{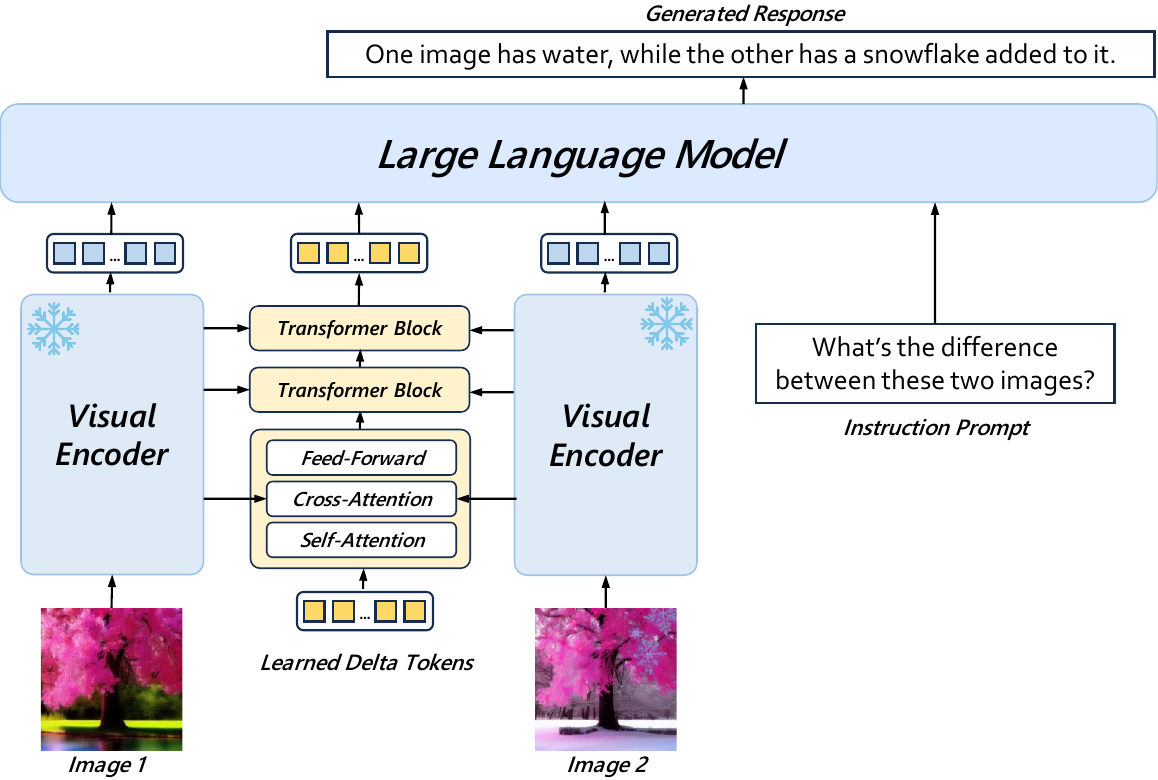}
    \caption{Architecture of our OneDiff model.}
    \label{fig:arch}
\end{figure}

\subsection{Model Architecture}

Our architecture leverages pretrained image encoders with designed modules and a cross-modal projector before a LLM to align image pairs with their correlative difference captions. We inherit excellent image representation ability from ready vision encoders, integrate them to derive preferable image features, and add delta tokens from the Visual Delta Module to promote capturing of subtle visual differences. The implemented vision encoder is CLIP-ViT-L/14 \cite{radford2021learning}, which can be generalized to any other vision encoder by our method. 

% \paragraph{Mixed Vision Encoders}

% To fully utilize the robust feature representation capabilities of vision encoders, this module effectively leverages multiple off-the-shelf encoders to obtain more comprehensive image features. We empirically select specific layer outputs from different encoders and amalgamate these features before aligning them with the LLM. In particular, we carefully choose hidden states of images from involved encoders, pass these hidden states through respective MLPs, and combine the features in a concatenating way before mapping to the semantic space of language modality. 

% This approach allows us to tap into the distinct strengths of various encoders, enhancing the depth and breadth of the feature representation obtained. By integrating these varied features, we aim to achieve a more comprehensive understanding and representation of the visual information, thereby enriching the overall performance and efficacy of our model. 

\paragraph{Visual Delta Module.}

To extract detailed features at multiple levels and discern the differences between image pairs, we have devised the Visual Delta Module, which actively engages with the hidden states of image pairs at regular intervals across several layers. This strategic design enables us to delve deep into the nuances of image features and highlight the distinctions between paired images. 

We employ a Transformer for interaction between vision encoders of two images. Specifically, a fixed number of learnable embeddings as so called delta tokens, which are utilized and input to the Transformer consisting of self-attention layers and cross-attention layers alternately, with image hidden states extracted from the encoder layers at equal intervals. These embeddings function as queries for image features in cross-attention layers, thus updated separately and simultaneously with pairs of image information. Our method can be conveniently adapted for models projecting encoded image features into the textual space.

The Visual Delta Module is intricately crafted to introduce a set number of delta tokens, strategically placed to capture fine-grained features perspective of the differences between image pairs. By incorporating these delta tokens, we aim to enhance the granularity of our feature extraction process, thereby enriching our understanding of the subtle variations between images within a pair.

\subsection{Model Training}

To endow models with capability to handle dual-image inputs and produce difference caption, we adopt a two-stage training strategy following LLaVA. For each stage, well-designed strategy and datasets are exploited to advance model performances. 

% \paragraph{Stage I: Alignment Stage with CST}
\paragraph{Stage I: Cross-Modal Alignment with Coupled Sample Training (CST).}
This initial stage functions as pretraining to effectively align visual and linguistic modalities. Utilizing high-quality image-text pairs from the LLaVA-558K dataset, we tailor our pretraining efforts to enhance alignment between image features and the language model. A key distinction in our approach is the simultaneous processing of paired images—unlike conventional VLMs that typically handle single images. This adaptation is crucial for maintaining consistency with the training objectives of IDC tasks.

Given the scarcity of datasets featuring paired images aligned with textual descriptions, we introduce the CST strategy to foster efficient training. CST ingeniously crafts dual-image samples from existing image-text pairs, effectively simulating the IDC scenario. This is achieved by dynamically merging two distinct image-text samples into a pseudo dual-image sample, where the model receives concatenated images as inputs and concatenated captions as outputs. This innovative setup challenges the model to synchronize complex image-text alignments and adapt to diverse contextual discrepancies.

During this stage, we focus on optimizing the Visual Delta Module and the projector for fine-grained inter-image features and cross-modal alignment. Meanwhile, the parameters of the visual encoders and the language model are kept frozen.

\paragraph{Stage II: Multi-Task Instruction Tuning with DiffCap Dataset}
In this stage, we leverage a multi-task instruction tuning  approach similar to typical VLMs. Here, each type of data within the DiffCap Dataset is presented to the model with task-specific prompts that guide the training process. This step ensures that the model not only learns general features from the training data but also hones its skills on specific types of differences that are critical for accurate IDC.

Specifically, our method introduce generated instruction tuning IDC data with multiple image inputs to mixed instruction-following data \cite{llava} including visual question answering, visual grounding, language conversation, etc. We implement IDC data generated from realistic paired images with upper-level differences and synthetic paired images with delicate differences. Collected datasets related to differences in image pairs vary from video-surveillance footage, editing images to bird photographs. Compared to the first training phase, the LLM in our architecture is unfrozen to preferably learn the representations of delta tokens output by the Visual Delta Module, and facilitate the extraction of cross-modal semantic information through interactions of images features. 

\paragraph{}
During our two-stage training process, the weights of vision encoders are preserved and fixed, considering that visual parts of the model exhibit powerful image representation capabilities and are enhanced by our meticulously designed Visual Delta Module. 

\section{Experiments}
We implement our OneDiff model for image difference captioning. Initially, we discuss the architectural design and training specifications. Subsequently, we detail the datasets employed for training and testing dual-image comprehension, along with their evaluation protocols. We then benchmark our model against current state-of-the-art counterparts. Lastly, we conduct an ablation study on the critical design elements of our model.

\subsection{Implementation Details}

\paragraph{Architecture.}
 The architecture incorporates a vision encoder, CLIP-ViT-L/14, a two-layer multilayer perceptron (MLP) projector, and the large language model (LLM) Vicuna-7B \cite{vicuna2023}. We utilize the penultimate layer outputs of the CLIP vision encoder for image feature extraction. This encoder consists of 24 layers. From each 336×336 image, we extracted 576 patches of size 24×24. Each patch possesses a dimensionality of 2048, which are subsequently mapped to 4096 to align within the LLM. In the Visual Delta Module (VDM) module, the input learnable delta tokens are initialized using a Gaussian distribution with a mean of 0 and a standard deviation of 0.02. This module is inherited from pretrained BERT-base-uncased \cite{devlin2018bert}. It consists of six layers featuring alternating self-attention and cross-attention blocks. We select image features from the 8th, 16th, and 24th layers of the vision encoder, facilitating interaction among them through delta tokens within the cross-attention blocks.

\paragraph{Training and Hyper-Parameters.}
% using the PyTorch framework
Our model is trained on 8 NVIDIA A800 (80G) GPUs for 10 hours (Vicuna) or 4 hours (Qwen2 \cite{yang2024qwen2}). During the pre-training phase, the initial learning rate is set at 1e-3 for MLP and 3e-4 for VDM, with a 3\% warm-up using a cosine decay schedule. Each GPU processes batches of 32. In the subsequent Supervised Fine-Tuning (SFT) phase, both the learning rate and batch size are reduced to 2e-5 and 8 per GPU, respectively. During pre-training, only the VDM and MLP modules are unfrozen, whereas in the SFT phase, a full fine-tuning of all parameters is implemented.

\paragraph{Training Data.} % 558 + 279 CST
In the pre-training phase, we employ LLaVA 558k \cite{llava} and 279k pairwise image captions generated by CST strategy to construct the pre-training data. In the SFT phase, we initially collect the currently available data, encompassing the training splits of three aforementioned benchmarks, CLEVR-Change \cite{johnson2017clevr} and NLVR2 \cite{suhr2018corpus}, termed the \textit{Coll} set. This set, along with both real and synthetic data obtained through our data generation pipeline, forms 335k image-text pairs designed for difference caption training, \textit{i.e.} DiffCap. Furthermore, to preserve the foundational visua-language comprehension capabilities of the model, we also incorporate LLaVA-665k, resulting in a comprehensive dataset of 1M pairs for instruction fine-tuning. Specifically, CLEVR-Change \cite{johnson2017clevr}, constructed automatically using the CLEVR engine, catalogs geometric object transformations against a plain background. The "after" images capture changes related to color, texture, addition, removal, and movement, alongside minimal viewpoint alterations. NLVR2 \cite{suhr2018corpus} evaluates whether paired visual data supports the statements. We randomly select a subset of NLVR2, retaining its original reasoning framework, and enhance it by adding instructions to generate data conducive to instruction-following tasks. For "True" statements, we refine them to include a new instruction: "Make a statement about these two images."

\subsection{Evaluation Benchmarks and Metrics}

We conduct experiments primarily on three benchmarks: Spot-the-Diff \cite{jhamtani2018learning}, Image-Editing-Request \cite{tan2019expressing}, and Birds-to-Words \cite{forbes2019neural}, which are described in further detail below. Consistent with prior research, we evaluate our models using BLEU (B) \cite{papineni2002bleu}, METEOR (M) \cite{banerjee2005meteor}, CIDEr (C) \cite{vedantam2015cider}, and ROUGE-L (R) \cite{lin2004rouge} as metrics. Specifically, Spot-the-Diff \cite{jhamtani2018learning} dataset is the inaugural dataset designed for the task of image difference captioning. It comprises 13,192 precisely aligned images sourced from surveillance cameras, characterized by a consistent viewpoint. Each image pair is annotated with a concise description of all significant differences, averaging 1.86 captions per pair. Image-Editing-Request \cite{tan2019expressing} dataset, sourced from 3,939 real image pairs derived from online image-editing forums, includes 5,695 editing directives. Typically, the modified objects in these images are subtle and indistinct. Birds-to-Words \cite{forbes2019neural} dataset documents the intricate distinctions among various bird species gathered in natural settings. Annotators focus solely on the visual differences between birds without referencing specific bird categories or the environmental background. This approach ensures that annotations are both precise and relevant to the observed phenotypic variations.

\begin{table}[tb]
    \caption{Comparison to SOTA on three IDC benchmarks based on BLEU-4 (B), CIDEr (C), METEOR (M), and ROUGE-L (R). }
    \begin{subtable}[t]{0.505\linewidth}
        \caption{Comparison on Image-Editing-Request.}
        \resizebox{\linewidth}{!}{ 
        \begin{tabular}{l|cccc} 
        \toprule
        \multirow{2}{*}{Model} & \multicolumn{4}{c}{Image-Editing-Request} \\
         % & BLEU-4 & CIDEr & METEOR & ROUGE-L\\
         & B & C & M & R\\
        \midrule
        Dyn rel-att \cite{tan2019expressing} & 6.7 & 26.4 & 12.8 & 37.3\\
        BiDiff \cite{sun2022bidirectional} & 6.9 & 27.7 & 14.6 & 38.5\\
        CLIP4IDC \cite{guo2022clip4idc} & 8.2 & 32.2 & 14.6 & 40.4\\
        NCT \cite{tu2023neighborhood} & 8.1 & 34.2 & 15.0 & 38.8\\
        VARD-LSTM \cite{tu2023adaptive} & 7.8 & 29.6 & 13.2 & 39.2\\
        VARD-Transformer \cite{tu2023adaptive} & 10.0 & 35.7 & 14.8 & 39.0\\
        SCORER \cite{tu2023self} & 9.6 & 31.0 & 14.6 & 39.5\\
        SCORER+CBR \cite{tu2023self} & 10.0 & 33.4 & 15.0 & 39.6\\
        \midrule
        OneDiff (ours) & \textbf{29.6} & \textbf{109.6} & \textbf{25.1} & \textbf{55.6}\\
        \bottomrule
        \end{tabular}
        }
        \label{sota:ier}
    \end{subtable}
    \hfill
    \begin{subtable}[t]{0.495\linewidth}
        \caption{Comparison on Spot-the-Diff.}
        \resizebox{\linewidth}{!}{
        \begin{tabular}{l|cccc} 
        \toprule
        \multirow{2}{*}{Model} & \multicolumn{4}{c}{Spot-the-Diff} \\
         % & BLEU-4 & CIDEr & METEOR & ROUGE-L\\
         & B & C & M & R\\
        \midrule
        M-VAM+RAF \cite{shi2020finding} & 11.1 & 42.5 & 12.9 & 33.2 \\
        VACC \cite{kim2021agnostic} & 9.7 & 41.5 & 12.6 & 32.1 \\
        CC-Full & 8.3 & 33.0 & 13.0 & 30.2 \\
        CLIP4IDC \cite{guo2022clip4idc} & 11.6 & 47.4 & 14.2 & 35.0 \\
        VARD-LSTM \cite{tu2023adaptive} & - & 39.3 & 13.1 & 33.1 \\
        VARD-Transformer\cite{tu2023adaptive} & - & 30.3 & 12.5 & 29.3\\
        SCORER \cite{tu2023self} & 9.4 & 38.5 & 13. & -\\
        SCORER+CBR \cite{tu2023self} & 10.2 & 38.9 & 12.2 & - \\
        \midrule
        OneDiff (ours) & \textbf{12.8} &\textbf{ 56.6} & \textbf{14.6} & \textbf{35.8} \\
        \bottomrule
        \end{tabular}
        }
        \label{sota:std}
    \end{subtable}

    \centering
    \begin{subtable}[t]{0.45\linewidth}
        \centering
        \caption{Comparison on Birds-to-Words.}
        \resizebox{\linewidth}{!}{ 
        \begin{tabular}{l|cccc} 
        \toprule
        \multirow{2}{*}{Model} & \multicolumn{4}{c}{Birds-to-Words} \\
         % & BLEU-4 & CIDEr & METEOR & ROUGE-L\\
         & B & C & M & R\\
         \midrule
        Neural Naturalist \cite{forbes2019neural} & 22.0 & 25.0 & -- & 43.0\\
        L2C \cite{yan2021l2c} & \textbf{31.8} & 16.3 & -- & 45.6\\
        IDC-PCL \cite{yao2022image} & 28.0 & 18.6 & -- & 48.4\\
         \midrule
        OneDiff (ours) & 27.1 & \textbf{28.4 }& \textbf{25.8 }& \textbf{49.6}\\
        \bottomrule
        \end{tabular}
        }
        \label{sota:btw}
    \end{subtable}

    \label{sota:all}
\end{table}

\subsection{Performance Comparison}

\begin{figure*}[tb]
    \centering
    \includegraphics[width=0.7\linewidth]{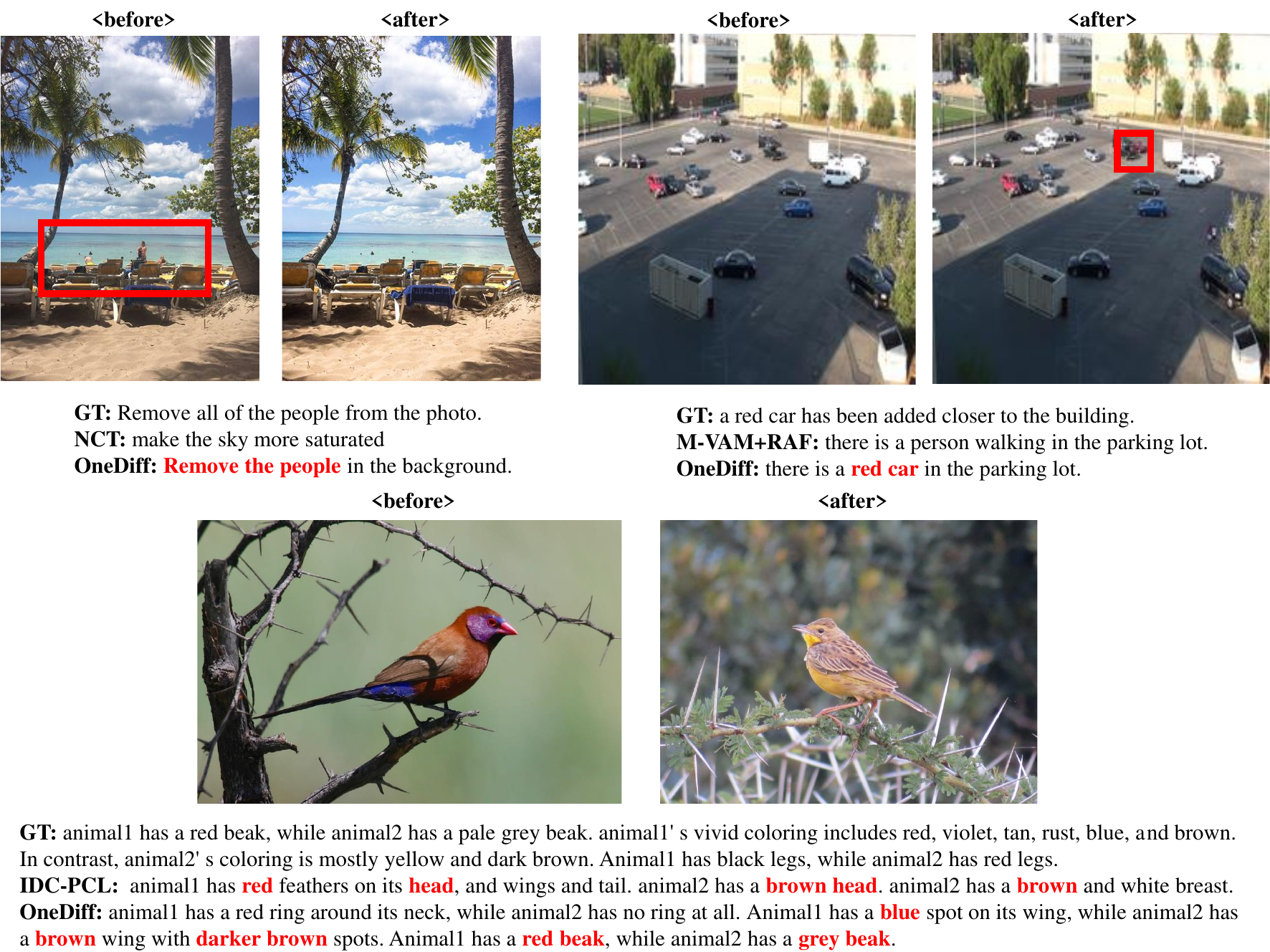}
    \caption{Examples of OneDiff generated results.}
    \label{fig:compare}
\end{figure*}

As shown in \ref{sota:ier}, we evaluate the performance of our model on Image-Editing-Request. OneDiff demonstrates significant superiority in this benchmark. Specifically, compared to the SOTA models, the improvement of our OneDiff on CIDEr is nearly doubled, \textit{i.e.} 211\% compare with SCORER+CBR. We then compare the performance on Spot-the-Diff. As detailed in \ref{sota:std}, OneDiff achieves significant improvements across BLEU, CIDEr and METEOR compared with SOTAs.
Finally, we analyze the performance disparities on the Birds-to-Words. Consistent with the previous two benchmarks, our model demonstrated a significant lead as shown in \ref{sota:btw}. Quantitative results comparing to other IDC methods are displayed in in \ref{fig:compare}. OneDiff generates more reasonable difference captions with correct details on the bad cases of these methods. 

% Performances of OneDiff on other single-image vision-language tasks are also explored in \ref{sota:sin}. Since OneDiff only involves dual-image inputs in Stage II, we select a subset from LLaVA-558K in addition to our DiffCap to achieve the comparable data volume with other large visual language models (LVLMs) for instruct tuning.

\begin{table*}[t]
\caption{Ablation of model design, including Coupled Sample Training (CST) strategy and Visual Delta Module (VDM). We conduct ablation analysis on three aforementioned benchmarks.}
\centering
\begin{tabular}{cc|cccc|cccc|cccc} 
\toprule
 \multirow{2}{*}{CST} & \multirow{2}{*}{VDM} & \multicolumn{4}{c}{Image-Editing-Request} & \multicolumn{4}{c}{Spot-the-Diff} & \multicolumn{4}{c}{Birds-to-Words} \\
 & & B & C & M & R & B & C & M & R & B & C & M & R\\
 \midrule
 × & × & 23.7 & 85.2 & 22.0 & 47.6 & 9.0 & 42.0 & 13.2 & 31.8 & 22.4 & 18.5 & 24.8 & 46.1\\
 \checkmark & × & 24.5 & 102.3 & 23.5 & 52.1 & 10.7 & 48.6 & 14.1 & 33.6 & 25.5 & 23.0 & 25.7 & 46.4\\
 × & \checkmark & 26.1 & 91.8 & 23.0 & 49.8 & 10.8 & 46.1 & 13.6 & 32.4 & 24.1 & 22.1 & 25.3 & 45.9\\
 \checkmark & \checkmark & 29.6 & 109.6 & 25.1 & 55.6 & 12.8 & 56.6 & 14.6 & 35.8 & 27.1 & 28.4 & 25.8 & 49.6\\
\bottomrule
\end{tabular}
\label{abl:1}
\end{table*}

\begin{table}[tb]
    \caption{Ablation of language models and Visual Delta Tokens on Birds-to-Words. }
    \begin{subtable}[t]{0.37\linewidth}
        \caption{A more efficient language model.}
        \resizebox{\linewidth}{!}{ 
        \begin{tabular}{l|cccc} 
        \toprule
        LLM & B & C & M & R \\
        \midrule
        Vicuna-7B & 25.0 & 21.8 & 25.9 & 46.7\\
        Qwen2-1.5B & 23.8 & 20.1 & 25.1 & 46.7\\
        \bottomrule
        \end{tabular}
        }
        \label{abl:llm}
    \end{subtable}
    \hfill
    \begin{subtable}[t]{0.63\linewidth}
        \caption{Impact of dropping tokens on performance of CIDEr.}
        \resizebox{\linewidth}{!}{
        \begin{tabular}{l|cccc} 
        \toprule
        Used Delta Tokens  & 100\% & 50\% & 20\% & 0\% \\
        \midrule
        OneDiff (Vicuna-7B) & 21.8 & 20.8 & 19.6 & 18.1\\
        \bottomrule
        \end{tabular}
        }
        \label{abl:vdt}
    \end{subtable}
    \label{abl:all}
\end{table}

\begin{table*}[t]
\caption{The effect of DiffCap data. Here, Coll, Real, and Syn represents the data components of the collected data, the generated data with real images, and the generated data with synthetic images, respectively.}
\centering
\begin{tabular}{l|cccc|cccc|cccc} 
\toprule
 \multirow{2}{*}{Data Components} & \multicolumn{4}{c}{Image-Editing-Request} & \multicolumn{4}{c}{Spot-the-Diff} & \multicolumn{4}{c}{Birds-to-Words} \\
 & B & C & M & R & B & C & M & R & B & C & M & R\\
  \midrule
 Coll & 22.3 & 90.1 & 22.0 & 51.5 & 8.9 & 40.7 & 13.8 & 31.8 & 23.7 & 20.8 & 24.4 & 46.9\\
 Coll+Syn & 26.5 & 94.4 & 24.0 & 51.5 & 9.0 & 42.0 & 14.3 & 32.8 & 25.5 & 21.8 & 25.5 & 46.7\\
 Coll+Real & 29.5 & 96.8 & 24.3 & 52.4 & 10.5 & 44.7 & 14.6 & 33.4 & 26.5 & 22.0 & 26.4 & 47.5\\
 Coll+Syn+Real & 29.6 & 109.6 & 25.1 & 55.6 & 12.8 & 56.6 & 14.6 & 35.8 & 27.1 & 28.4 & 25.8 & 49.6\\
 % & Coll+Real+Syn+LLaVA-288K & 28.4 & 108.1 & 24.7 & 55.2 & 9.8 & 48.5 & 12.7 & 31.8 & 27.1 & 18.8 & 26.2 & 47.2\\
\bottomrule
\end{tabular}
\label{abl:2}
\end{table*}

\subsection{Ablation Analysis}

\paragraph{Ablation of Model Design.}
We first assess the efficacy of the proposed CST strategy and VDM module. We report the related results in \ref{abl:1} with the 6-layer and 32-delta-token VDM. The hyper-parameter setting of VDM is discussed in our supplementary materials. Specifically, on one hand, the CST strategy facilitates cost-effective and efficient adaptation for model to multi-image understanding domains, yielding improvements across all benchmarks. 
On the other hand, while the isolated implementation of the VDM module does not significantly outperform the baseline (which we attribute to insufficient training of the VDM due to the CST being disabled during the pretrain phase), the concurrent activation of both CST and VDM substantially boosts performance demonstrating the synergistic benefits, significantly surpassing the performance of a single strategy.
We opt for a more efficient language model in \ref{abl:llm}, reducing computational costs while maintaining commendable performance on Birds-to-Words. To further demonstrate the role of VDT, we randomly drop a portion of them during inference to test the impact on performance on Birds-to-Words in \ref{abl:vdt}. 

% proposed methods. 

\paragraph{The Effect of DiffCap Data.}

In this section, we systematically analyze the impact of both realistic and synthetic data generated via our proposed pipeline, as indicated by the performance metrics outlined in \ref{abl:2}.  Specifically, incorporating either type of data independently yield consistent performance gains across three benchmarks, which attests to the high quality and extensive distribution of our data. Moreover, utilizing both data types in tandem further enhance performance, underscoring the complementary nature of the two data patterns. Overall, the data we constructed have substantially augmented the existing IDC dataset, effectively mitigating the data scarcity issue in this field.

% \paragraph{Qualitative Analysis.}
% We show the qualitative examples of generated results in Figure \ref{fig:example}.

\section{Conclusion}
In this paper, we introduce the OneDiff model, a novel generalist approach to Image Difference Captioning (IDC) that effectively utilizes a siamese image encoder and a Visual Delta Module within a robust vision-language framework. We also present the DiffCap dataset, a comprehensive, large-scale collection of IDC data that integrates various existing IDC datasets alongside our newly generated real-world and synthetic images, facilitated by an automated generation pipeline. Trained with our innovative Coupled Sample Training strategy and the DiffCap dataset, OneDiff significantly enhances IDC capabilities across diverse contexts without necessitating task-specific adjustments. Our rigorous evaluations on benchmarks such as Spot-the-Diff, CLEVR-Change, and Birds-to-Words highlight the model's exceptional performance, establishing new standards for flexibility and efficiency in IDC. Furthermore, our work underscores the potential of large-scale visual language models to discern exceedingly subtle image differences and changes, suggesting promising applications in fields such as forgery detection, medical image analysis, and aerial change detection.

\begin{credits}
\subsubsection{\ackname} This research is supported by Artificial Intelligence-National Science and Technology Major Project (2023ZD0121200), the National Natural Science Foundation of China (No. 62436001, 62437001).
\end{credits}
%%%%%%%%%%%%%%%%%%%%%%%%%%%%%%%%%%%%%%%%%%%%%%%%%%%%%%%%%%%%%%%%%%%%%%%%%%%%%%%%%%%%%%%%%%%%%%%%%%%%%%%

% \clearpage\mbox{}Page \thepage\ of the manuscript.
% \clearpage\mbox{}Page \thepage\ of the manuscript.
% \clearpage\mbox{}Page \thepage\ of the manuscript.
% \clearpage\mbox{}Page \thepage\ of the manuscript.
% \clearpage\mbox{}Page \thepage\ of the manuscript. This is the last page.
% Now we have reached the maximum length of an ACCV \ACCVyear{} submission (excluding references).
% References should start immediately after the main text, but can continue past p.\ 14 if needed.
% \clearpage  % TODO REVIEW/FINAL: This \clearpage needs to be removed from both review and camera-ready versions.

\bibliographystyle{splncs04}
\bibliography{main}

\begin{thebibliography}{10}
\providecommand{\url}[1]{\texttt{#1}}
\providecommand{\urlprefix}{URL }
\providecommand{\doi}[1]{https://doi.org/#1}

\bibitem{achiam2023gpt}
Achiam, J., Adler, S., Agarwal, S., Ahmad, L., Akkaya, I., Aleman, F.L., Almeida, D., Altenschmidt, J., Altman, S., Anadkat, S., et~al.: Gpt-4 technical report. arXiv preprint arXiv:2303.08774  (2023)

\bibitem{alayrac2022flamingo}
Alayrac, J.B., Donahue, J., Luc, P., Miech, A., Barr, I., Hasson, Y., Lenc, K., Mensch, A., Millican, K., Reynolds, M., et~al.: Flamingo: a visual language model for few-shot learning. Advances in neural information processing systems  \textbf{35},  23716--23736 (2022)

\bibitem{antol2015vqa}
Antol, S., Agrawal, A., Lu, J., Mitchell, M., Batra, D., Zitnick, C.L., Parikh, D.: Vqa: Visual question answering. In: Proceedings of the IEEE international conference on computer vision. pp. 2425--2433 (2015)

\bibitem{banerjee2005meteor}
Banerjee, S., Lavie, A.: Meteor: An automatic metric for mt evaluation with improved correlation with human judgments. In: Proceedings of the acl workshop on intrinsic and extrinsic evaluation measures for machine translation and/or summarization. pp. 65--72 (2005)

\bibitem{brooks2023instructpix2pix}
Brooks, T., Holynski, A., Efros, A.A.: Instructpix2pix: Learning to follow image editing instructions. In: Proceedings of the IEEE/CVF Conference on Computer Vision and Pattern Recognition. pp. 18392--18402 (2023)

\bibitem{brown2020language}
Brown, T.B.: Language models are few-shot learners. arXiv preprint arXiv:2005.14165  (2020)

\bibitem{vicuna2023}
Chiang, W.L., Li, Z., Lin, Z., Sheng, Y., Wu, Z., Zhang, H., Zheng, L., Zhuang, S., Zhuang, Y., Gonzalez, J.E., Stoica, I., Xing, E.P.: Vicuna: An open-source chatbot impressing gpt-4 with 90\%* chatgpt quality (March 2023), \url{https://lmsys.org/blog/2023-03-30-vicuna/}

\bibitem{devlin2018bert}
Devlin, J., Chang, M.W., Lee, K., Toutanova, K.: Bert: Pre-training of deep bidirectional transformers for language understanding. arXiv preprint arXiv:1810.04805  (2018)

\bibitem{forbes2019neural}
Forbes, M., Kaeser-Chen, C., Sharma, P., Belongie, S.: Neural naturalist: Generating fine-grained image comparisons. arXiv preprint arXiv:1909.04101  (2019)

\bibitem{guo2022clip4idc}
Guo, Z., Wang, T.J.J., Laaksonen, J.: Clip4idc: Clip for image difference captioning. arXiv preprint arXiv:2206.00629  (2022)

\bibitem{hudson2019gqa}
Hudson, D.A., Manning, C.D.: Gqa: A new dataset for real-world visual reasoning and compositional question answering. In: Proceedings of the IEEE/CVF conference on computer vision and pattern recognition. pp. 6700--6709 (2019)

\bibitem{jhamtani2018learning}
Jhamtani, H., Berg-Kirkpatrick, T.: Learning to describe differences between pairs of similar images. arXiv preprint arXiv:1808.10584  (2018)

\bibitem{johnson2017clevr}
Johnson, J., Hariharan, B., Van Der~Maaten, L., Fei-Fei, L., Lawrence~Zitnick, C., Girshick, R.: Clevr: A diagnostic dataset for compositional language and elementary visual reasoning. In: Proceedings of the IEEE conference on computer vision and pattern recognition. pp. 2901--2910 (2017)

\bibitem{kim2021agnostic}
Kim, H., Kim, J., Lee, H., Park, H., Kim, G.: Agnostic change captioning with cycle consistency. In: Proceedings of the IEEE/CVF International Conference on Computer Vision. pp. 2095--2104 (2021)

\bibitem{li2023blip}
Li, J., Li, D., Savarese, S., Hoi, S.: Blip-2: Bootstrapping language-image pre-training with frozen image encoders and large language models. In: International conference on machine learning. pp. 19730--19742. PMLR (2023)

\bibitem{lin2004rouge}
Lin, C.Y.: Rouge: A package for automatic evaluation of summaries. In: Text summarization branches out. pp. 74--81 (2004)

\bibitem{lin2014microsoft}
Lin, T.Y., Maire, M., Belongie, S., Hays, J., Perona, P., Ramanan, D., Doll{\'a}r, P., Zitnick, C.L.: Microsoft coco: Common objects in context. In: Computer Vision--ECCV 2014: 13th European Conference, Zurich, Switzerland, September 6-12, 2014, Proceedings, Part V 13. pp. 740--755. Springer (2014)

\bibitem{llava}
Liu, H., Li, C., Wu, Q., Lee, Y.J.: Visual instruction tuning (2023)

\bibitem{papineni2002bleu}
Papineni, K., Roukos, S., Ward, T., Zhu, W.J.: Bleu: a method for automatic evaluation of machine translation. In: Proceedings of the 40th annual meeting of the Association for Computational Linguistics. pp. 311--318 (2002)

\bibitem{park2019robust}
Park, D.H., Darrell, T., Rohrbach, A.: Robust change captioning. In: Proceedings of the IEEE/CVF International Conference on Computer Vision. pp. 4624--4633 (2019)

\bibitem{plummer2015flickr30k}
Plummer, B.A., Wang, L., Cervantes, C.M., Caicedo, J.C., Hockenmaier, J., Lazebnik, S.: Flickr30k entities: Collecting region-to-phrase correspondences for richer image-to-sentence models. In: Proceedings of the IEEE international conference on computer vision. pp. 2641--2649 (2015)

\bibitem{radford2021learning}
Radford, A., Kim, J.W., Hallacy, C., Ramesh, A., Goh, G., Agarwal, S., Sastry, G., Askell, A., Mishkin, P., Clark, J., et~al.: Learning transferable visual models from natural language supervision. In: ICML (2021)

\bibitem{sak2014long}
Sak, H., Senior, A., Beaufays, F.: Long short-term memory based recurrent neural network architectures for large vocabulary speech recognition. arXiv preprint arXiv:1402.1128  (2014)

\bibitem{shi2020finding}
Shi, X., Yang, X., Gu, J., Joty, S., Cai, J.: Finding it at another side: A viewpoint-adapted matching encoder for change captioning. In: Computer Vision--ECCV 2020: 16th European Conference, Glasgow, UK, August 23--28, 2020, Proceedings, Part XIV 16. pp. 574--590. Springer (2020)

\bibitem{suhr2018corpus}
Suhr, A., Zhou, S., Zhang, A., Zhang, I., Bai, H., Artzi, Y.: A corpus for reasoning about natural language grounded in photographs. arXiv preprint arXiv:1811.00491  (2018)

\bibitem{sun2022bidirectional}
Sun, Y., Li, L., Yao, T., Lu, T., Zheng, B., Yan, C., Zhang, H., Bao, Y., Ding, G., Slabaugh, G.: Bidirectional difference locating and semantic consistency reasoning for change captioning. International Journal of Intelligent Systems  \textbf{37}(5),  2969--2987 (2022)

\bibitem{tan2019expressing}
Tan, H., Dernoncourt, F., Lin, Z., Bui, T., Bansal, M.: Expressing visual relationships via language. arXiv preprint arXiv:1906.07689  (2019)

\bibitem{10562229}
Tang, Y., Wang, W., Zhang, C., Liu, J., Zhao, Y.: Learnable feature augmentation framework for temporal action localization. IEEE Transactions on Image Processing  \textbf{33},  4002--4015 (2024). \doi{10.1109/TIP.2024.3413599}

\bibitem{10183357}
Tang, Y., Wang, W., Zhang, C., Liu, J., Zhao, Y.: Temporal action proposal generation with action frequency adaptive network. IEEE Transactions on Multimedia  \textbf{26},  2340--2353 (2024). \doi{10.1109/TMM.2023.3295090}

\bibitem{team2023gemini}
Team, G., Anil, R., Borgeaud, S., Wu, Y., Alayrac, J.B., Yu, J., Soricut, R., Schalkwyk, J., Dai, A.M., Hauth, A., et~al.: Gemini: a family of highly capable multimodal models. arXiv preprint arXiv:2312.11805  (2023)

\bibitem{tu2023adaptive}
Tu, Y., Li, L., Su, L., Du, J., Lu, K., Huang, Q.: Adaptive representation disentanglement network for change captioning. IEEE Transactions on Image Processing  (2023)

\bibitem{tu2023neighborhood}
Tu, Y., Li, L., Su, L., Lu, K., Huang, Q.: Neighborhood contrastive transformer for change captioning. IEEE Transactions on Multimedia  (2023)

\bibitem{tu2023self}
Tu, Y., Li, L., Su, L., Zha, Z.J., Yan, C., Huang, Q.: Self-supervised cross-view representation reconstruction for change captioning. In: Proceedings of the IEEE/CVF International Conference on Computer Vision. pp. 2805--2815 (2023)

\bibitem{vedantam2015cider}
Vedantam, R., Lawrence~Zitnick, C., Parikh, D.: Cider: Consensus-based image description evaluation. In: Proceedings of the IEEE conference on computer vision and pattern recognition. pp. 4566--4575 (2015)

\bibitem{wang2023cogvlm}
Wang, W., Lv, Q., Yu, W., Hong, W., Qi, J., Wang, Y., Ji, J., Yang, Z., Zhao, L., Song, X., et~al.: Cogvlm: Visual expert for pretrained language models. arXiv preprint arXiv:2311.03079  (2023)

\bibitem{yan2021l2c}
Yan, A., Wang, X.E., Fu, T.J., Wang, W.Y.: L2c: Describing visual differences needs semantic understanding of individuals. arXiv preprint arXiv:2102.01860  (2021)

\bibitem{yang2024qwen2}
Yang, A., Yang, B., Hui, B., Zheng, B., Yu, B., Zhou, C., Li, C., Li, C., Liu, D., Huang, F., et~al.: Qwen2 technical report. arXiv preprint arXiv:2407.10671  (2024)

\bibitem{yao2022image}
Yao, L., Wang, W., Jin, Q.: Image difference captioning with pre-training and contrastive learning. In: Proceedings of the AAAI Conference on Artificial Intelligence. vol.~36, pp. 3108--3116 (2022)

\bibitem{zhang2023llama}
Zhang, R., Han, J., Liu, C., Gao, P., Zhou, A., Hu, X., Yan, S., Lu, P., Li, H., Qiao, Y.: Llama-adapter: Efficient fine-tuning of language models with zero-init attention. arXiv preprint arXiv:2303.16199  (2023)

\end{thebibliography}

\appendix
\section{Appendix}

In this appendix section, we provide following items:
\begin{itemize}
% \item \ref{a1} Ablation experiments on the language model in our \textbf{OneDiff} framework and the hyper-parameters settings of the VDM module.
\item \ref{a1} Ablation experiments onthe hyper-parameters settings of the VDM module in our \textbf{OneDiff} framework.

% \item \ref{a2} Further exploration on performances of \textbf{OneDiff} on single-image vision-language tasks.
% \item \ref{a2} Further exploration on performances of \textbf{OneDiff} on single-image vision-language tasks and zero-shot performances on IDC tasks.

\item \ref{a3} The statistics of our proposed \textbf{DiffCap Dataset}, both real-image and synthetic-image samples from DiffCap and some potential failure cases. 
\end{itemize}

\subsection{Ablation}
\label{a1}
The hyper-parameter setting of VDM is explored in \ref{abl:4}. For efficiency, we only train 3000 steps in Stage II when exploring the model design. The number of delta tokens varies in 16, 32, 64 and 128 with 6 or 12 Transformer layers. OneDiff is not quite sensitive to different hyper-parameters with relatively stable performances. 

\begin{table*}[htb]
\caption{Ablation of model design on Visual Delta Module (VDM). We conduct ablation analysis on three aforementioned benchmarks.}
\centering
\begin{tabular}{cc|cccc|cccc|cccc} 
\toprule
 \multirow{2}{*}{Layers} & \multirow{2}{*}{Tokens} & \multicolumn{4}{c}{Image-Editing-Request} & \multicolumn{4}{c}{Spot-the-Diff} & \multicolumn{4}{c}{Birds-to-Words} \\
 & & B & C & M & R & B & C & M & R & B & C & M & R\\
 \midrule
 6 & 16 & 28.1 & 98.8 & 24.0 & 51.0 & 9.8 & 43.1 & 13.6 & 32.4 & 25.0 & 19.6 & 26.0 & 46.4\\
6 & 32 & 28.4 & 99.7 & 23.4 & 51.4 & 10.1 & 47.6 & 13.8 & 32.4 & 25.9 & 20.6 & 25.9 & 46.8\\
 6 & 64 & 28.7 & 100.9 & 24.0 & 52.0 & 10.5 & 45.5 & 14.3 & 33.1 & 24.4 & 18.3 & 25.1 & 46.2\\
6 & 128 & 26.0 & 93.1 & 22.7 & 49.7 & 10.1 & 45.7 & 13.5 & 32.4 & 24.7 & 19.5 & 25.9 & 45.9\\
 12 & 16 & 22.9 & 94.7 & 23.0 & 50.8 & 9.6 & 42.8 & 13.8 & 32.1 & 26.1 & 21.8 & 25.9 & 46.3\\
 12 & 32 & 23.4 & 99.9 & 23.5 & 50.7 & 9.2 & 40.1 & 13.2 & 31.6 & 24.9 & 22.2 & 25.7 & 46.6\\
 12 & 64 & 27.4 & 97.6 & 23.6 & 50.3 & 9.8 & 44.5 & 13.6 & 32.5 & 23.9 & 16.6 & 25.8 & 45.0\\
 12 & 128 & 24.0 & 101.6 & 23.9 & 52.1 & 9.7 & 44.4 & 13.9 & 32.4 & 24.8 & 20.6 & 26.0 & 46.9\\
\bottomrule
\end{tabular}
\label{abl:4}
\end{table*}

\subsection{Data}
\label{a3}

This section illustrates the statistics of DiffCap and the qualitative results of data construction.

\subsubsection{Data Statistics.}

Our DiffCap dataset consists of three components: collected datasets, generated data from real images and generated data from synthetic images. (See \ref{data:diff})

\begin{table}[htb]
\caption{Statistics of our DiffCap dataset. "Real" refers to our generated data from real images, and "Synthetic" refers to our generated data from synthetic images.}
\centering
\begin{tabular}{llrr} 
\toprule
Component & Sub-Component & \#Image-Pairs & \#Annotations \\
 \midrule
\multirow{5}{*}{Collected} & Spot-the-Diff & 11K & 21K\\
 & CLEVR-Change & 8K & 44K\\
 & Image-Editing-Request & 3.4K & 4.2K\\
 & Birds-to-Words & 14K & 14K\\
 & NLVR2 & 86K & 86K\\
\midrule
\multirow{2}{*}{Generated} & Real & 54K & 54K\\
 & Synthetic & 139K & 139K\\
\midrule
Total &  & 316K & 363K\\
\bottomrule
\end{tabular}
\label{data:diff}
\end{table}

\subsubsection{Data Samples.} 
\begin{figure}[htb]
    \centering
    \includegraphics[width=\linewidth]{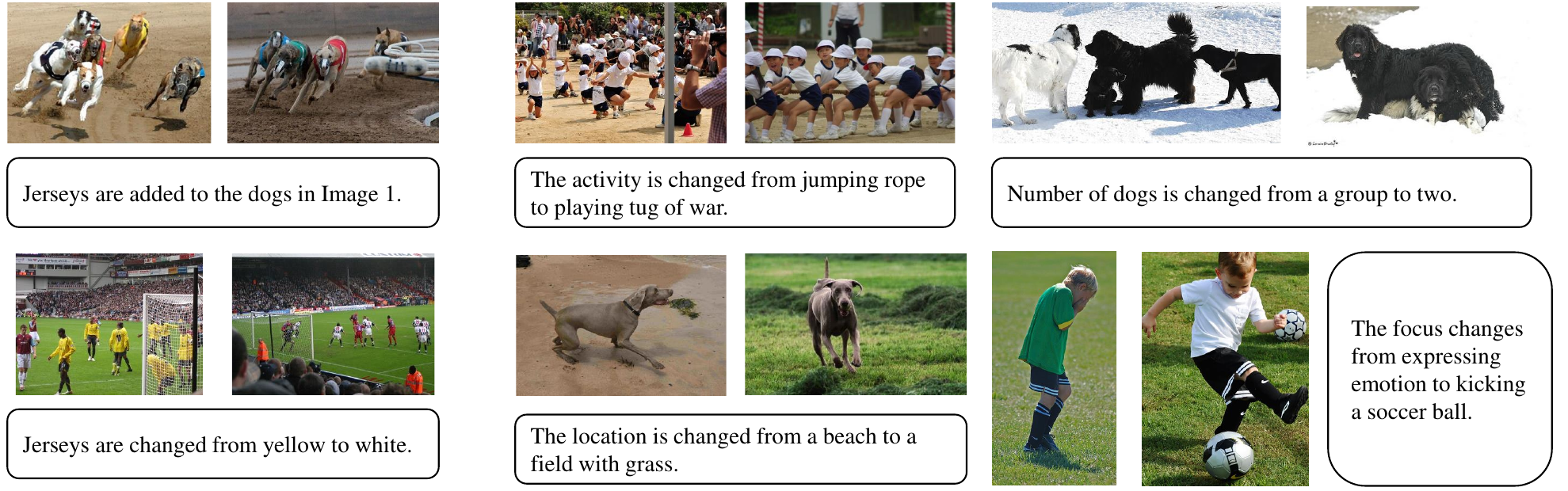}
    \caption{More real-image samples of our DiffCap dataset.}
    \label{fig:sam}
\end{figure}

\begin{figure}[htb]
    \centering
    \includegraphics[width=\linewidth]{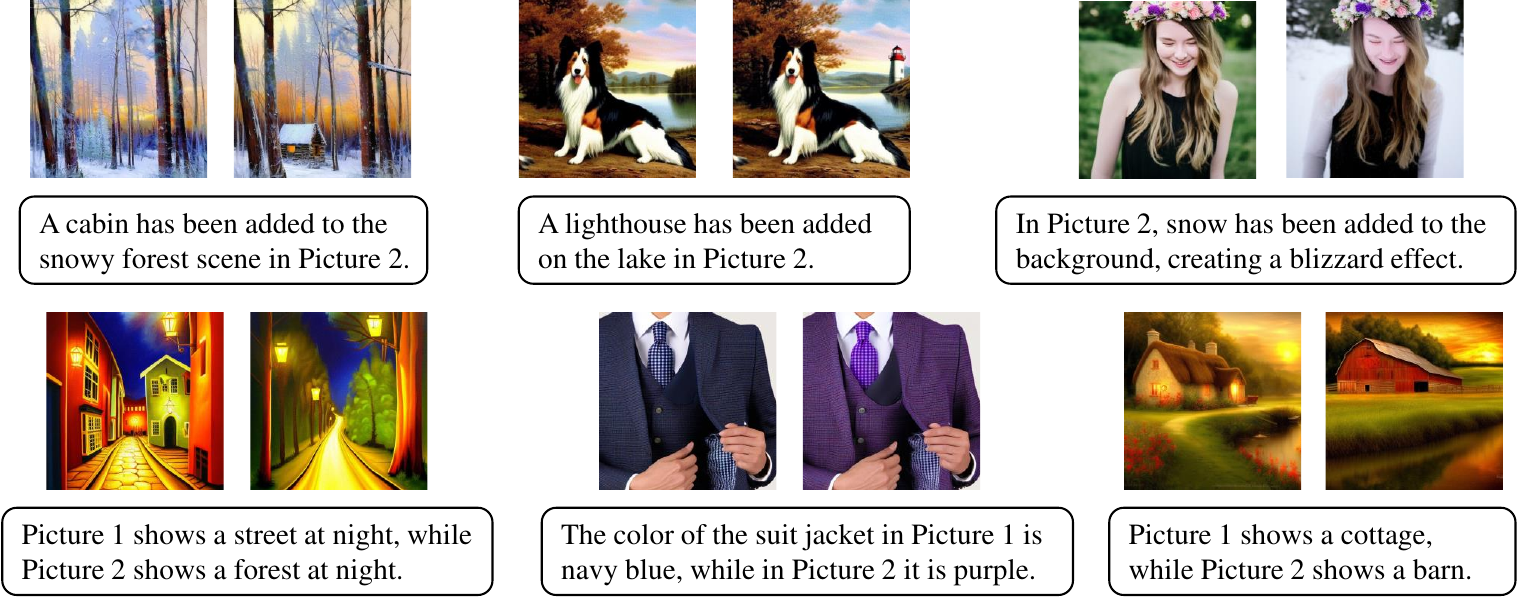}
    \caption{More synthetic-image samples of our DiffCap dataset.}
    \label{fig:sam2}
\end{figure}

More samples from DiffCap are displayed as qualitative results of date construction. (See \ref{fig:sam} and \ref{fig:sam2})

\subsubsection{Potential Failure Cases} 
\begin{figure}[htb]
    \centering
    \includegraphics[width=\linewidth]{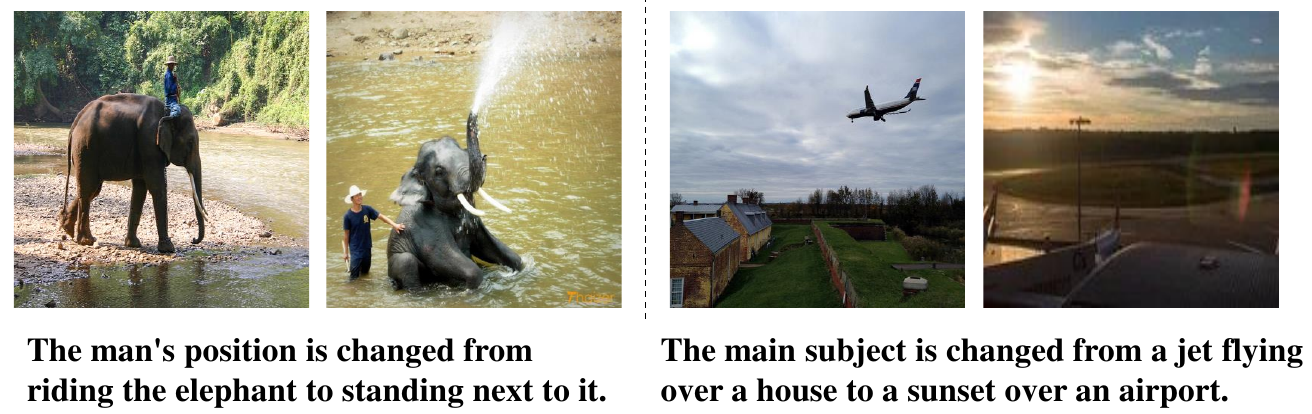}
    \caption{Potential failure cases of our DiffCap dataset.}
    \label{fig:bad}
\end{figure}

We present potential failure cases where the environment and objectives are roughly the same, yet multiple describable image differences exist, which could lead to confusion. (See \ref{fig:bad})

\end{document}